\documentclass[letterpaper]{article} % DO NOT CHANGE THIS
\usepackage{aaai2026}  % DO NOT CHANGE THIS
\usepackage{times}  % DO NOT CHANGE THIS
\usepackage{helvet}  % DO NOT CHANGE THIS
\usepackage{courier}  % DO NOT CHANGE THIS
\usepackage[hyphens]{url}  % DO NOT CHANGE THIS
\usepackage{graphicx} % DO NOT CHANGE THIS
\usepackage{natbib}  % DO NOT CHANGE THIS AND DO NOT ADD ANY OPTIONS TO IT
\usepackage{caption} % DO NOT CHANGE THIS AND DO NOT ADD ANY OPTIONS TO IT
\usepackage{algorithm}
\usepackage{algorithmic}
\usepackage{subcaption}
\usepackage{enumitem}
\newcounter{checksubsection}
\newcounter{checkitem}[checksubsection]
\usepackage{xcolor}

\usepackage{newfloat}
\usepackage{listings}
\usepackage{multirow}
\usepackage{amsfonts}
\usepackage{amsmath}
\usepackage{booktabs}
\usepackage{dashrule}
\DeclareCaptionStyle{ruled}{labelfont=normalfont,labelsep=colon,strut=off} % DO NOT CHANGE THIS
\floatstyle{ruled}
\newfloat{listing}{tb}{lst}{}
\floatname{listing}{Listing}
\title{DR.Experts: Differential Refinement of Distortion-Aware Experts for \\ Blind Image Quality Assessment}
\author {
    Bohan Fu\textsuperscript{\rm 1\thanks{Equal contribution, \newline\hangindent=1.5em$\dagger$ Corresponding author.}},
	Guanyi Qin\textsuperscript{\rm 2\footnotemark[1]},
    Fazhan Zhang\textsuperscript{\rm 1}, 
    Zihao Huang\textsuperscript{\rm 1},
    Mingxuan Li\textsuperscript{\rm 1},
	Runze Hu\textsuperscript{\rm 1$\dagger$},
}

\affiliations {
 \textsuperscript{\rm 1} Beijing Institute of Technology, Beijing 100086, China\\
 \textsuperscript{\rm 2} National University of Singapore, Singapore 119276, Singapore\\
 \{fubohan809, zfz63622, huangzihhhh, limx1630, hrzlpk2015\}@gmail.com, guanyi.qin@u.nus.edu
}

\begin{document}

\maketitle

\begin{abstract}
Blind Image Quality Assessment, aiming to replicate human perception of visual quality without reference, plays a key role in vision tasks, yet existing models often fail to effectively capture subtle distortion cues, leading to a misalignment with human subjective judgments. We identify that the root cause of this limitation lies in the lack of reliable distortion priors, as methods typically learn shallow relationships between unified image features and quality scores, resulting in their insensitive nature to distortions and thus limiting their performance. To address this, we introduce DR.Experts, a novel prior-driven BIQA framework designed to explicitly incorporate distortion priors, enabling a reliable quality assessment. DR.Experts begins by leveraging a degradation-aware vision-language model to obtain distortion-specific priors, which are further refined and enhanced by the proposed Distortion-Saliency Differential Module through distinguishing them from semantic attentions, thereby ensuring the genuine representations of distortions. The refined priors, along with semantics and bridging representation, are then fused by a proposed mixture-of-experts style module named the Dynamic Distortion Weighting Module. This mechanism  weights each distortion-specific feature as per its perceptual impact, ensuring that the final quality prediction aligns with human perception. Extensive experiments conducted on five challenging BIQA benchmarks demonstrate the superiority of DR.Experts over current methods and showcase its excellence in terms of generalization and data efficiency. 

\end{abstract}

\begin{links}
    \link{Code}{https://github.com/FuBohan01/DR.Experts}
\end{links}

\section{Introduction}

\begin{figure}[!ht]
  \centering
  \hspace{-2mm}\includegraphics[width=0.9\linewidth]{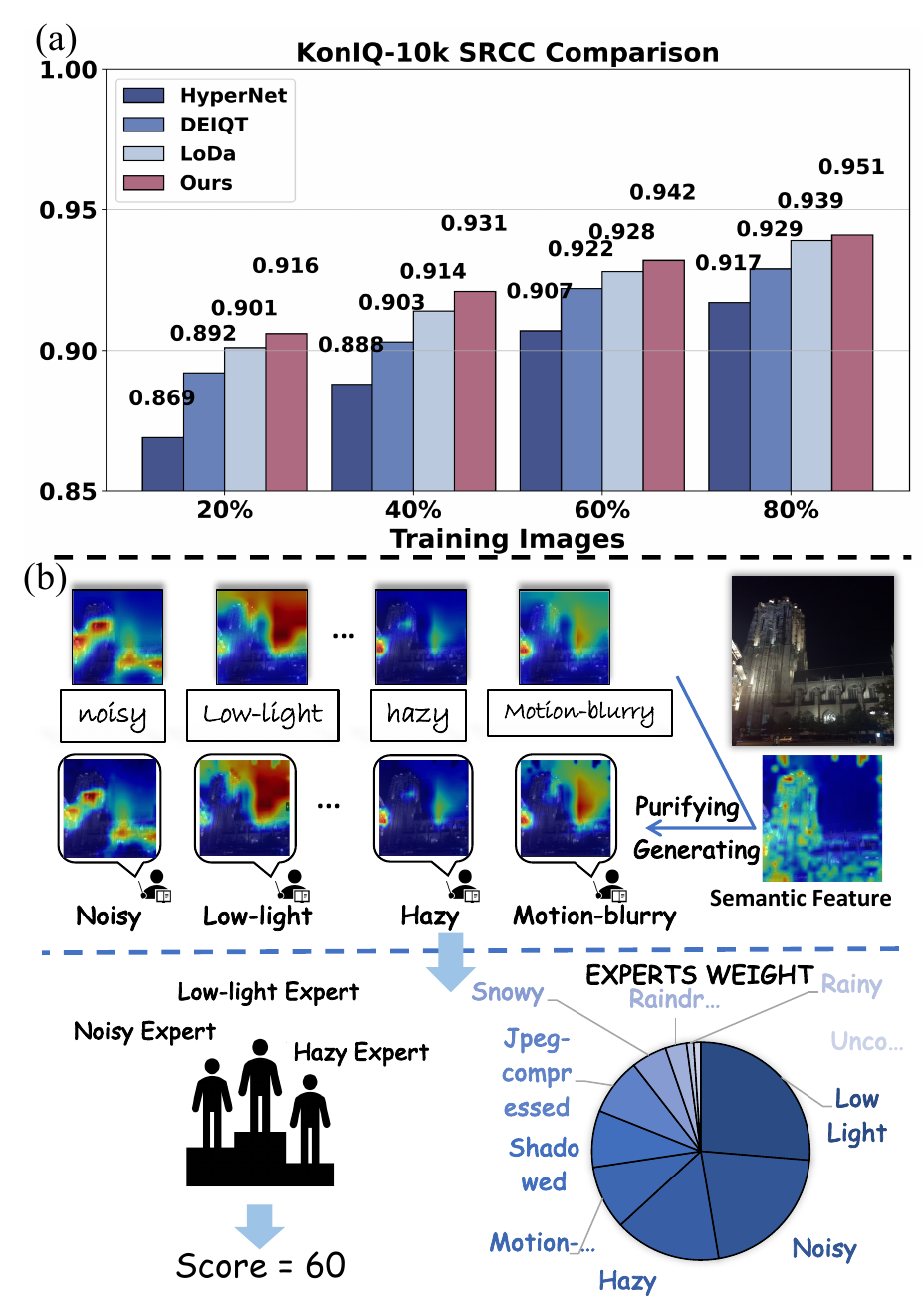}
  \caption{(a): DR.Experts maintains the advantage even with limited training data. (b): Given a distorted image, the proposed framework DR.Experts first leverages a vision-language model specialized on visual distortions to obtain attention corresponding to various distortions. By differentiating these cues from semantic attention, DR.Experts effectively further purifies distortion-aware representations. These refined features are then adaptively weighted according to their perceptual importance and integrated to yield a precise and perceptually consistent quality assessment.
  }
  \label{fig:demo}
\end{figure}
% \vspace{0cm}

High-quality images are essential for reliable performance in vision tasks. Being instructional and serving as a crucial and valuable part in enabling quality control in real-world scenarios, \textbf{Blind Image Quality Assessment} (BIQA) aims to evaluate image visual quality without reference images, easing the burden of references that are hard to acquire in in-the-wild cases~\cite{huang2024scale, li2025few, 6272356}. Albeit numerous efforts have been made to push the boundaries of BIQA, the advantage of being reference-free still introduces significant performance challenges, especially in terms of distortion modeling~\cite{moorthy2011blind}. Existing BIQA techniques still struggle to effectively characterize quality-aware representations, undermining their abilities to align with human evaluations of perceptual quality.

A primary ground of such limitations arises from their reduced sensitivity to subtle distortions and conceptual ambiguity under complicated and diverse distortions due to the lack of guidance from distortion priors. While BIQA has made significant progress~\cite{chen2024topiq} through the development of intricate learning architectures and advanced training strategies recently, these methods~\cite{yang2024align} still focus on learning a plain relationship between consolidated and unified image features and the final quality scores. However, in-the-wild images often suffer from diverse types and varying levels of distortions~\cite{hosu2020koniq}, such as under-exposure, noise, and motion blur, affecting image visual quality in different ways and extents. Yet, the limited volume of existing BIQA datasets, coupled with a lack of distortion specifications and annotations, constrains the capabilities of models to learn and accurately capture these subtle yet important distortion-related features.

As the challenge in capturing distortion-specific characteristics highlights a fundamental gap in existing BIQA approaches, thus, naturally, a key motivation to advance BIQA is to effectively incorporate prior knowledge of distortions into current learning frameworks~\cite{li2024bridging}, enabling fine-grained and perceptually aligned modeling of quality-aware representations. Driven by this insight, we propose a \textbf{D}ifferential \textbf{R}inement of Distortion-Aware \textbf{Experts} BIQA framework, hereby named \textbf{DR.Experts}, that incorporates distortion priors to guide the extraction of perceptual visual quality-relevant features, and distill and purify these features further by differentiating them with semantic features. These refined features are then aggregated with adaptive weights as per their respective contributions to the image quality, acting as multiple experts for different distortion perspectives, thus yielding a comprehensive and trustworthy quality prediction, as shown in Fig.~\ref{fig:demo}. Such a design not only enhances the alignment between prediction and human perception but also offers transparency in the decision-making process, as the generated score can be traced back to certain distortion factors, thereby increasing the trustworthiness of the assessment.

To this end, CLIP~\cite{radford2021learning} is incorporated into the proposed framework for distortion priors. Serving as a foundation between visuals and text, Contrastive Language-Image Pre-Training (CLIP) enables transferable priors by mapping both modalities into a shared embedding space, where text can prompt distortion-related visual features. Specifically, we employ DA-CLIP~\cite{luo2024controlling}, a derivative optimized for low-level vision tasks and fine-tuned to increase sensitivity to distortion features. DA-CLIP demonstrates strong capabilities for fine-grained distortion perception, it achieves an impressive average accuracy of $99.2\%$ on tasks involving ten distinct distortion types, including blur, low light, and compression errors. Based on this, we obtain the distortion-aware visual attentions activated by DA-CLIP under various distortion-specific prompts as priors by the dot product between the prompt representations from the text encoder and the visual features from the image encoder. To further strengthen these representations and suppress the attention noise of semantic redundancy originating from upstream classification pre-training of CLIP and Vision Transformer (ViT)~\cite{dosovitskiy2020image}, we introduce the \textbf{Distortion-Saliency Differential Module} (DSDM), with its \textbf{differential refinement attention mechanism} to isolate distortion-related features. DSDM refines the priors by contrasting DA-CLIP's distortion-aware attention with the semantic attention extracted by ViT, thereby ensuring a precise quality assessment.

Furthermore, to effectively integrate and leverage features of different types, along with considerations that various distortions impact perceptual quality in different ways and extents, we design the \textbf{Dynamic Distortion Weighting Module} (DDWM), a Mixture-of-Experts (MoE) architecture that adaptively aggregates these feature groups: distortion priors refined by DSDM, semantic features extracted by ViT, and bridging features as intermediate supplementaries derived from the difference between the two aforementioned feature groups. This module not only model complementary cues across diverse representations, and allows to to assign adaptive weights to different distortion types based on their respective perceptual impacts. By doing so, our proposed framework aligns more closely with the fine-grained assessment criteria of the human vision system, further enhancing both the accuracy and trustworthiness of the generated quality score. To summarize, our contributions can be regarded as follows:
\begin{itemize}
    \item We propose a novel BIQA framework that leverages distortion priors as guidance, enabling fine-grained, distortion-aware quality assessment. By leveraging DA-CLIP’s text-prompted transferable visual attention as priors, our framework learns towards trustworthy and explainable results. The proposed framework enhances the perceptual alignment between human-perceived distortion representations and quality metrics.
    \item A novel module called Distortion-Saliency Differential Module is introduced to refine distortion priors by differentiating DA-CLIP’s distortion-aware attention with ViT-derived semantic attention. DSDM effectively suppresses suppressing redundant semantic noise from pre-training, while enhances the saliency of distortion features.
    \item Dynamic Distortion Weighting Module is further proposed to dynamically assign significance weights to score token regarding different types and levels of distortions, aiming at simulating a comprehensive analysis as per subject quality assessment of distortion characteristics by specialists.
    \item We verify the proposed framework on five diverse and challenging BIQA benchmarks, where it consistently outperforms other competitors and showcases strong generalization and data efficiency capabilities.
\end{itemize}

\section{Related Work}
\subsection{Blind Image Quality Assessment}
Conventional BIQA methods mainly use hand-crafted features based on natural scene statistics~\cite{zhang2015feature, wang2021associations, 6172573}.  However, due to the constrained representation resulting from manual feature selection, these methods exhibit limited generalization in in-the-wild scenes. With the advancement of deep learning, BIQA has seen notable improvements. Current learning-based approaches can be categorized into CNN-based and ViT-based methods~\cite{chu2025attention}, extracting image features for straightforward end-to-end regression of quality scores~\cite{8576582, kang2014convolutional, bosse2017deep}. By increasing capacity and depth, models, \emph{e.g.}, ResNet~\cite{he2016deep} and ViT~\cite{dosovitskiy2020image}, pre-trained on large datasets~\cite{deng2009imagenet} are used to further perceive image distortion~\cite{shin2024blind, xu2023local}. To overcome the restricted volume of IQA datasets, some self-supervised learning methods conduct contrastive learning on degraded images formed by distortion models~\cite{agnolucci2024arniqa, saha2023re}, to fully perceive the differences and similarities of distortions. At the same time, domain adaptation, multi-scale feature adaptation, and other techniques~\cite{chu2025union, hu2025toward, Qin_2025_ICCV} have also been explored to advance BIQA~\cite{su2020blindly}. 
Recently, multimodal large language models have been widely applied in downstream vision tasks ~\cite{wu2024towards, li2023blip}, with researchers also exploring their powerful visual understanding and perception capabilities on BIQA by fine-tuning, and thus to promot the interpretability and robustness of BIQA research~\cite{ht_mm, wu2024qa}.
\subsection{CLIP for Low-level Vision}
Built upon the foundations of large-scale (image, text) pairs and contrastive pre-training, CLIPs~\cite{radford2021learning}, a series of models, demonstrate strong zero-shot capabilities for transfer learning and have thus been widely applied in downstream vision tasks~\cite{agnolucci2024reference, liang2023iterative}. CLIP-IQA, serving as the first piece of work, explores the possibility of using CLIP for BIQA through a sequence of carefully designed prompts~\cite{wang2023exploring}. Meanwhile, methods such as multitask learning and self-supervised strategies have been further employed to leverage the knowledge of CLIP in the field of BIQA.~\cite{agnolucci2024quality, kwon2024attiqa}. DA-CLIP adapted CLIP for low-level vision tasks by introducing a controller to accurately predict the distortion type of the input image, yielding excellent results of $99.2$ accuracy~\cite{luo2024controlling}. In this paper, we utilize the prior of specific distortion types obtained by DA-CLIP to guide our proposed model to assess image quality in a fine-grained manner.

\section{Methodology}

\subsection{Overview}
\begin{figure*}[!ht]
  \centering
\includegraphics[width=\linewidth]{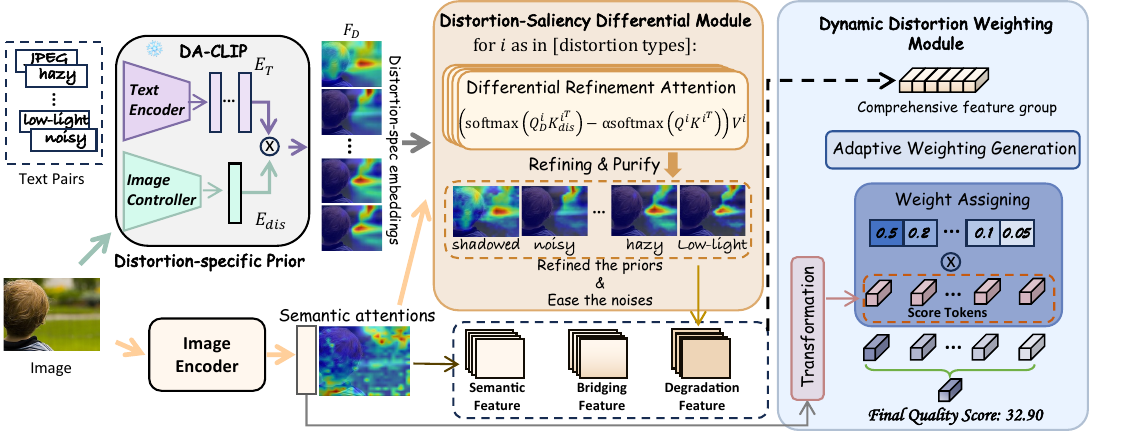}
  \caption{Overall architecture of the proposed DR.Experts. We leverage DA-CLIP to obtain priors and use DSDM to refine the attentions. DDWM then, serving as experts, to weigh the importance of distortions and give final predictions.}
  \label{fig:framework}
\end{figure*}

In this work, We propose a novel BIQA network that evaluates images precisely under the guidence of the prior of specific distortion types within the image to form a mixture-of-Experts. The overall framework is illustrated in the Fig.~\ref{fig:framework}. Specifically, an RGB image is used as input, and a Vision Transformer is employed as the Image Encoder. Meanwhile, DA-CLIP is utilized to identify the distortion types present in the image. Subsequently, we introduce a Distortion-Saliency Differential Module to refine different image distortion information by removing noise unrelated to distortion features and enhances the saliency of distortion features. To avoid the influence of non-dominant or absent distortion types in the image on the final quality assessment, we designed the Dynamic Distortion Weighting module to integrate the semantic feature from Image Encoder, distortion priors refined by DSDM, and supplementary interval features derived from the difference between them. The comprehensive feature group formed by the above three features assigns dynamic weights to experts of different distortion types and obtains the final image quality score.

\subsection{Distortion-specific Prior}

In most previous works, data-driven approaches have been heavily relied upon. However, we introduce prior knowledge of distortion types, enabling the model to extract features in a more targeted manner and thereby reducing excessive dependence on data-driven methods. In this work, we adopt DA-CLIP as the Distortion-specific Prior Module. Built upon the frozen CLIP text and Image Encoder, DA-CLIP trains an additional image controller through contrastive learning to predict high-quality image distortion feature embeddings.
CLIP is a multimodal model that aligns images and text in a shared embedding space, widely used in cross-modal retrieval and understanding tasks. DA-CLIP is designed to generate features that match the actual distortion type of the input image, making it better suited to the needs of degraded scenarios. Specifically, we use DA-CLIP's feature embeddings for image distortion types as prior knowledge for image quality assessment tasks.
DA-CLIP maps the input RGB image $I \in \mathbb{R}^{H \times W \times C}$ and ten predefined common image distortion types $T$ = [motion-blurry, hazy, jpeg-compressed, low-light, noisy, raindrop, rainy, shadowed, snowy, uncompleted], into the same feature space through the image controller $\mathcal{E_D}$ and text encoder $\mathcal{E_T}$, respectively. Specifically, the degraded image representation is expressed as $E_{dis} = \mathcal{E}_D(I)$, while the text representation for the $i$-th distortion type is given by $ E_T^i = \mathcal{E}_T(T^i)$. We use the Hadamard product similar to CLIPIQA~\cite{wang2023exploring} to obtain the representation of the image under different distortion categories. This process can be formulated as:
\begin{equation}
\begin{aligned}
    % &E_{degra} = \mathcal{E}_D(I), E_T^i = \mathcal{E}_T(T^i), \\
    &F_D^i = \text{Linear}^i\left( E_{dis} \odot E_T^i \right).
\end{aligned}
\end{equation}

\subsection{Distortion-Saliency Differential Module}
Inspired by the Differential Transformer~\cite{ye2025differential}, which effectively suppresses noise in homogenous attention through a differential mechanism, we extend this approach to heterogeneous attention and propose the Differential Refinement Attention Mechanism. This mechanism is capable of suppressing redundant or overlapping information, including attention noise and certain high-level semantic information, between distortion prior knowledge and the semantic information of the Image Encoder. Furthermore, it enhances the prior knowledge related to distortions. Specifically, an RGB image is processed by a Vision Transformer to extract high-level semantic features $F \in \mathbb{R}^{N \times E}$, where $N$ is the sequence length and $E$ is the embedding dimension. Simultaneously, the RGB image $I$ is processed by DA-CLIP to obtain features $F_D^i \in \mathbb{R}^{N \times E}$, where $i \in [1,2,...,10]$ ,corresponding to a specific distortion type. $F_D^i$, proposed as the degraded category information query in DA-CLIP, aims to extract features related to specific types of distortion from the image features $E_{dis}$ generated by the image controller of DA-CLIP. At this stage, the distortion features contain a small amount of semantic information and attention noise. The query and key for the semantic features extracted by Image Encoder are defined as $Q^i = W_Q^iF$ and $K^i = W_K^iF$, while the query and key for the features of $i$-th distortion type are defined as $Q_D^i = W_{DQ}^iF_D^i$ and $K_{dis}^i = W_{disK}^iE_{dis}$. Moreover, the value vector is computed as $V^i = W_V^i\text{Cat}([F, F_D^i])$. Using these, the differential refinement attention mechanism extracts the image features under the $i$-th distortion type as follows:
\begin{equation}
\begin{aligned}
    F_{distortion}^i = (\text{softmax}(Q_D^i{K_{dis}^i}^T) - \\
    \alpha \text{softmax}(Q^i{K^i}^T))V^i,
\end{aligned}
\end{equation}
where, $F_{distortion}^i$is the feature of the image under the $i$-th distortion type. $\alpha$ is a learnable parameter. As a result, the module isolates and refine the features that correspond exclusively to a specific distortion type.

Subsequently, the features of all distortion types are fed into a Feed-Forward Network (FFN) composed of two linear layers with GELU as the activation function, thereby generating expert knowledge for identifying different distortion features.
\begin{equation}
\begin{aligned}
    &F_{multi-distortion} = \text{FFN}(F_{distortion}^1, ... , F_{distortion}^i), \\
\end{aligned}
\end{equation}
where, $\text{FFN}(\cdot)$ is the Feed-Forward Network. Benefiting from the Differential Refinement Attention Mechanism, this module can effectively suppress redundant or overlapping information, accurately extract features related to different distortion types, and further integrate and optimize multi-distortion features through the FFN, thereby generating expert-level representations for image distortions.

\subsection{Dynamic Distortion Weighting Module}

To assign distortion weights based on the impact of image distortions on perceived image quality and thereby achieve more reliable quality assessment scores, we propose a hybrid expert system named the Dynamic Distortion Weighting Module (DDWM). 
This module first utilizes the semantic features $F$ extracted by the Image Encoder, and the distortion priors $F_{Group}=(1-\lambda)\cdot F_{multi-distortion} $, as refined by the DSDM with a learnable parameter $\lambda$ to balance the contributions to the following modeling. Supplementary interval features,  named by bridging feature $F_{bridging} = F- F_{Group}$ are further incorporated into this module to form a comprehensive feature group $F_{com}= \text{Cat}([F_{distortion}^i, F, F_{bridging}]$ for multidimensional quality evaluation. The supplementary interval features serve to bridge distributional differences between dimensions or levels of features, ensuring a more cohesive representation.

The supplementary interval features serve to bridge distributional differences between dimensions or levels of features, ensuring a more cohesive representation.

Based on this feature group, the Adaptive Weighting Generation (WG) module dynamically calculates the importance of each distortion type in terms of its impact on quality perception, \emph{i.e.}, the dynamic expert weight for each distortion type. This process could described as:
\begin{equation}
\begin{aligned}
    W_{distortion}^1,...,W_{distortion}^{10} = \text{WG}(F_{com}).
\end{aligned}
\end{equation}
Here, Weighting Generation refers to a multi-layer perceptron (MLP) with PReLU as the activation function.
Finally, the computed weights are combined with the token $T_{score}$ derived from the ViT’s class token through a linear transformation, representing the current image quality. The score weight is employed to optimize and adjust the final image quality score, which can be written as:
\begin{equation}
\begin{aligned}
    \text{Score} = \sum_{i=1}^{10} W_{distortion}^iT_{score}
\end{aligned}
\end{equation}

\section{Experiments}

\subsection{Benchmark Datasets}

We assess the performance of the proposed DR.Experts on five in-the-wild BIQA datasets collected from real-world scenarios. The BID~\cite{ciancio2010no} and LIVEC~\cite{ghadiyaram2015massive} datasets comprise 590 and 1,162 images taken by photographers using mobile devices, respectively. KonIQ-10k~\cite{hosu2020koniq} contains 10,073 images selected from publicly available multimedia sources. SPAQ~\cite{fang2020perceptual} includes 11,125 images captured with smartphone cameras, covering diverse content and resolutions. LIVEFB~\cite{ying2020patches}, the largest authentic dataset to date, consists of 39,810 images.

Two widely used BIQA metrics are adopted as performance evaluation metrics, including Spearman’s Rank-order Correlation Coefficient (SRCC) and Pearson’s Linear Correlation Coefficient (PLCC), to measure the effectiveness of DR.Experts. SRCC evaluates the monotonic relationship between predicted and ground-truth scores, while PLCC reflects the prediction accuracy of models. Both metrics range from $0$ to $1$, with higher values indicating better performance. Ideally, achieving values closer to $1$ demonstrates superior prediction monotonicity and accuracy.

\subsection{Implementation Details}

For our approach, we apply a standard data augmentation technique commonly used in BIQA, where each image is randomly cropped into smaller patches. In addition, we follow the same experimental settings as QPT~\cite{zhao2023quality} and QCN~\cite{shin2024blind}. The number of patches is adjusted based on the size of each dataset. The Image Encoder in our method is the small version of DeiT-III~\cite{touvron2022deit}, pre-trained on ImageNet with a weight decay of $0.05$, a batch size of 1024, and trained for 400 epochs. As for the prior knowledge module, we utilize the pre-trained DA-CLIP image controller and text encoder~\cite{luo2024controlling}. This module is frozen during training.

DR.Experts is then fine-tuned on BIQA datasets for $9$ epochs with the prior module frozen. The learning rate starts at $2\times 10^{-4}$ and is reduced by a factor of $10$ after every $3$ epochs. The Smooth L1 loss is utilized as the loss function for model training, and the batch size is determined by the scale of the datasets, \emph{e.g.}, $64$ for the LIVEC dataset and $156$ for the KonIQ dataset. For all datasets, $80\%$ of the images are splitted and used for training, while the remaining $20\%$ are reserved for testing. We repeat this train-test setting $10$ times to mitigate the splitting bias, and record the medians of the results. We use PyTorch to implement DR.Experts, and all experiments are conducted using $4$ RTX 4090 GPUs.

\subsection{Comparison with SOTAs}

In this experiment, 14 representative BIQA methods are involved and compared with DR.Experts in terms of prediction performance on the in-the-wild datasets. Among them, DIIVINE~\cite{zhang2015feature} and BRISQUE~\cite{6272356} are based on handcrafted perceptual features, while the remaining methods are based on learning architectures, \emph{e.g.}, DB-CNN~\cite{8576582} and HyperIQA~\cite{su2020blindly} utilize CNNs, while MUSIQ~\cite{ke2021musiq} and DEIQT~\cite{qin2023data} leverage vision transformers. Additionally, we included TReS~\cite{golestaneh2022no} and LODA~\cite{xu2024boosting}, methods that combine the strengths of both CNNs and ViT, as well as the LQmamba~\cite{guan2025qmamba} based on the vision mamba architecture.

\begin{table*}[!ht]
    \centering
    \resizebox{\textwidth}{!}{
        \begin{tabular}{lcccccccccc}
            \toprule[1.5pt]
             & \multicolumn{2}{c}{KonIQ} & \multicolumn{2}{c}{LIVEC} & \multicolumn{2}{c}{SPAQ} & \multicolumn{2}{c}{LIVEFB} & \multicolumn{2}{c}{BID} \\
            \cmidrule(lr){2-3} \cmidrule(lr){4-5} \cmidrule(lr){6-7} \cmidrule(lr){8-9} \cmidrule(lr){10-11}
             Methods & SRCC & PLCC & SRCC & PLCC & SRCC & PLCC & SRCC & PLCC & SRCC & PLCC \\
            \midrule
            ILNIQE~\cite{zhang2015feature}       & 0.503 & 0.496 & 0.453 & 0.511 & 0.719 & 0.654 & 0.219 & 0.255 & 0.495 & 0.454 \\
            BRISQUE~\cite{6272356}      & 0.715 & 0.702 & 0.601 & 0.621 & 0.802 & 0.806 & 0.320 & 0.356 & 0.574 & 0.540 \\
            WaDIQaM-NR~\cite{bosse2017deep}   & 0.729 & 0.754 & 0.692 & 0.730 & 0.840 & 0.845 & 0.435 & 0.430 & 0.653 & 0.636 \\
            DB-CNN~\cite{8576582}       & 0.878 & 0.887 & 0.844 & 0.862 & 0.910 & 0.913 & 0.554 & 0.652 & 0.845 & 0.850 \\
            HyperIQA~\cite{su2020blindly}     & 0.906 & 0.917 & 0.859 & 0.882 & 0.916 & 0.919 & 0.535 & 0.623 & 0.869 & 0.878 \\
            MUSIQ~\cite{ke2021musiq}        & 0.916 & 0.928 & 0.702 & 0.746    & 0.917 & 0.921 & --    & --    & 0.646 & 0.739 \\
            TReS~\cite{golestaneh2022no}         & 0.915 & 0.928 & 0.846 & 0.877 & --    & --    & 0.554 & 0.625 & --    & --    \\
            DEIQT~\cite{qin2023data}        & 0.921 & 0.934 & 0.875 & 0.894 & 0.919 & 0.923 & 0.571 & 0.663 & --    & --    \\
            \midrule
            CONRTIQUE$^\star$~\cite{madhusudana2022image}   & 0.894 & 0.906 & 0.845 & 0.857 & 0.914 & 0.919 & 0.580 & 0.641 & --    & --    \\
            QPT$^\star$~\cite{zhao2023quality}         & 0.927 & 0.941 & 0.895 & 0.914 & 0.925 & 0.928 & 0.578 & 0.675 & 0.888 & 0.911 \\
            QCN~\cite{shin2024blind}          & 0.934 & 0.945 & 0.875 & 0.893 & 0.923 & 0.928 & -- & -- & 0.892 & 0.890 \\
            QFM-IQM~\cite{li2024adaptive} & 0.922 & 0.936 & 0.891 & 0.913 & 0.920 & 0.924 & 0.567 & 0.667 & -- & -- \\
            LODA~\cite{xu2024boosting}  & 0.932 & 0.944 & 0.876 & 0.899 & 0.925 & 0.928 & 0.578 & 0.679 & -- & -- \\
            LQMamba~\cite{guan2025qmamba} & 0.928 & 0.943 & 0.863 & 0.903 & 0.927 & 0.933 & 0.574 & 0.672 & -- & -- \\
            \midrule[1pt]
            DR.Experts (Ours) & \textbf{0.941} & \textbf{0.954} & \textbf{0.914} & \textbf{0.926} & \textbf{0.928} & \textbf{0.933} & \textbf{0.585} & \textbf{0.690} & \textbf{0.896} & \textbf{0.919} \\
            \bottomrule[1.5pt]
        \end{tabular}
        }
    \caption{Performance comparison measured by medians of SRCC and PLCC, where bold entries indicate the top two results. Pre-training methods are marked with the $\star$.}
    \label{tab:iqa_results}
\end{table*}

As shown in Table~\ref{tab:iqa_results}, DR.Experts exhibits significant improvements across five diverse BIQA datasets involving a range of complex scenes and levels of distortion. Given that achieving robust performance on these datasets is inherently challenging due to their varied nature, and compared to the SOTA approaches such as QFM-IQM~\cite{li2024adaptive}, LODA, LQMamba and QCN~\cite{shin2024blind}, DR.Experts achieves at least a $0.75$- and a $0.95$-point improvement on the KonIQ-10k dataset in terms of the SRCC and PLCC metrics, respectively. Our design of leverage distortion priors are further confirm. Additionally, DR.Experts outperforms methods that leverage quality-pretraining techniques, such as QPT~\cite{zhao2023quality} and CONRTIQUE~\cite{madhusudana2022image}, further demonstrating its robustness and adaptability. These results firmly emphasize the capabilities of our DR.Experts to effectively utilize and refine distortion priors and characteristics, particularly in the context of challenging in-the-wild scenarios.

\subsection{Generalization Validation}

\begin{table}[!ht]
    \centering
    \resizebox{0.99\columnwidth}{!}{
    \begin{tabular}{lcccc}
        \toprule[1.5pt]
        TRAINING & \multicolumn{2}{c}{LIVEFB} & LIVEC & KonIQ \\
        \cmidrule(lr){1-1} \cmidrule(lr){2-3} \cmidrule(lr){4-4} \cmidrule(lr){5-5}
        TESTING & KonIQ & LIVEC & KonIQ & LIVEC \\
        \midrule
        DBCNN   & 0.716 & 0.724 & 0.754 & 0.755 \\
        P2P-BM  & 0.755 & 0.738 & 0.740 & 0.770 \\
        HperlQA & 0.758 & 0.735 & 0.772 & 0.785 \\
        TReS    & 0.713 & 0.740 & 0.733 & 0.786 \\
        DEIQT   & 0.733 & 0.781 & 0.744 & 0.794 \\
        LODA    & 0.763 & 0.805 & 0.745 & 0.811 \\
        \midrule[1pt]
        DR.Experts   & \textbf{0.783} & \textbf{0.807} & \textbf{0.785} & \textbf{0.841} \\
        Gains   & 0.020($\uparrow$) & 0.002($\uparrow$) & 0.040($\uparrow$) & 0.030($\uparrow$) \\
        \bottomrule[1.5pt]
    \end{tabular}
    }
    \caption{SRCC on the generalization validation. The best performance is highlighted in bold. Gains are calculated versus the second-best performance.}
    \label{tab:crossdataset}
\end{table}

We conducted a cross-dataset experiment to verify the generalization ability of the model, which is an essential indicator of BIQA practices, where a model is trained and tested on different datasets. The results are shown in Table~\ref{tab:crossdataset}. In short, across all different training and test pairs, DR.Experts has achieved better results in terms of the SRCC metric compared to previous SOTA methods. This result also verifies the core idea of leveraging prior knowledge of distortion to help enhance the robustness and generalization.

\subsection{Data Efficiency Validation}

Benefiting from priors, together with the quality-aware attention further refined through DSDM, our model could quickly attend to distortions and thus alleviate its performance reliance on large-scale data. To verify this, we conducted data efficiency validation by reducing the training set data to $60\%$, $40\%$, and $20\%$ of the original size, while maintaining the same model architecture and training settings, and compared it with existing models. As shown in Table~\ref{tab:data-efficiency}, under the $60\%$ mode, DR.Experts are $2.3$- and $3$-point higher than LoDa on the LIVEC dataset in terms of the SRCC and PLCC metrics, respectively. Notably, under the $20\%$ mode, DR.Experts surpasses with the largest performance improvement against other methods. Those results further confirm the good robustness and generalization ability of DR.Experts and its core idea of priors and refinement.

\begin{table}[!ht]
    \centering
    \resizebox{0.99\columnwidth}{!}{
    \begin{tabular}{llcccc}
        \toprule[1.5pt]
        & & \multicolumn{2}{c}{KonIQ} & \multicolumn{2}{c}{LIVEC} \\ 
        \cmidrule(lr){3-4} \cmidrule(lr){5-6}
         Mode & Methods & SRCC & PLCC & SRCC & PLCC \\ 
        \midrule[1pt]
        \multirow{4}{*}{\large{20$\%$}} 
         & HyperNet & 0.869 & 0.873 & 0.776 & 0.809 \\
         & DEIQT & 0.888 & 0.908 & 0.792 & 0.822 \\
         & LoDa & 0.907 & 0.923 & 0.815 & 0.854 \\
         & DR.Experts & \textbf{0.917} & \textbf{0.931} & \textbf{0.837} & \textbf{0.861} \\   
        \midrule[1pt]
        \multirow{4}{*}{\large{40$\%$}} 
         & HyperNet & 0.892 & 0.908 & 0.832 & 0.849 \\
         & DEIQT & 0.903 & 0.922 & 0.838 & 0.855 \\
         & LoDa & 0.922 & 0.935 & 0.849 & 0.879 \\
         & DR.Experts & \textbf{0.929} & \textbf{0.942} & \textbf{0.874} & \textbf{0.896} \\   
        \midrule[1pt]
        \multirow{4}{*}{\large{60$\%$}} 
         & HyperNet & 0.901 & 0.914 & 0.843 & 0.862 \\
         & DEIQT & 0.914 & 0.931 & 0.848 & 0.877 \\
         & LoDa & 0.928 & 0.940 & 0.869 & 0.891 \\
         & DR.Experts & \textbf{0.939} & \textbf{0.950} & \textbf{0.899} & \textbf{0.914} \\   
        \bottomrule[1.5pt]
    \end{tabular}
    }
    \caption{Data-efficient learning validation with the training set containing $20\%$, $40\%$ and $60\%$ images.  Bold entries indicate the best performance.}
    \label{tab:data-efficiency}
\end{table}

\begin{figure*}[!ht]
  \centering
  \includegraphics[width=0.9\linewidth]{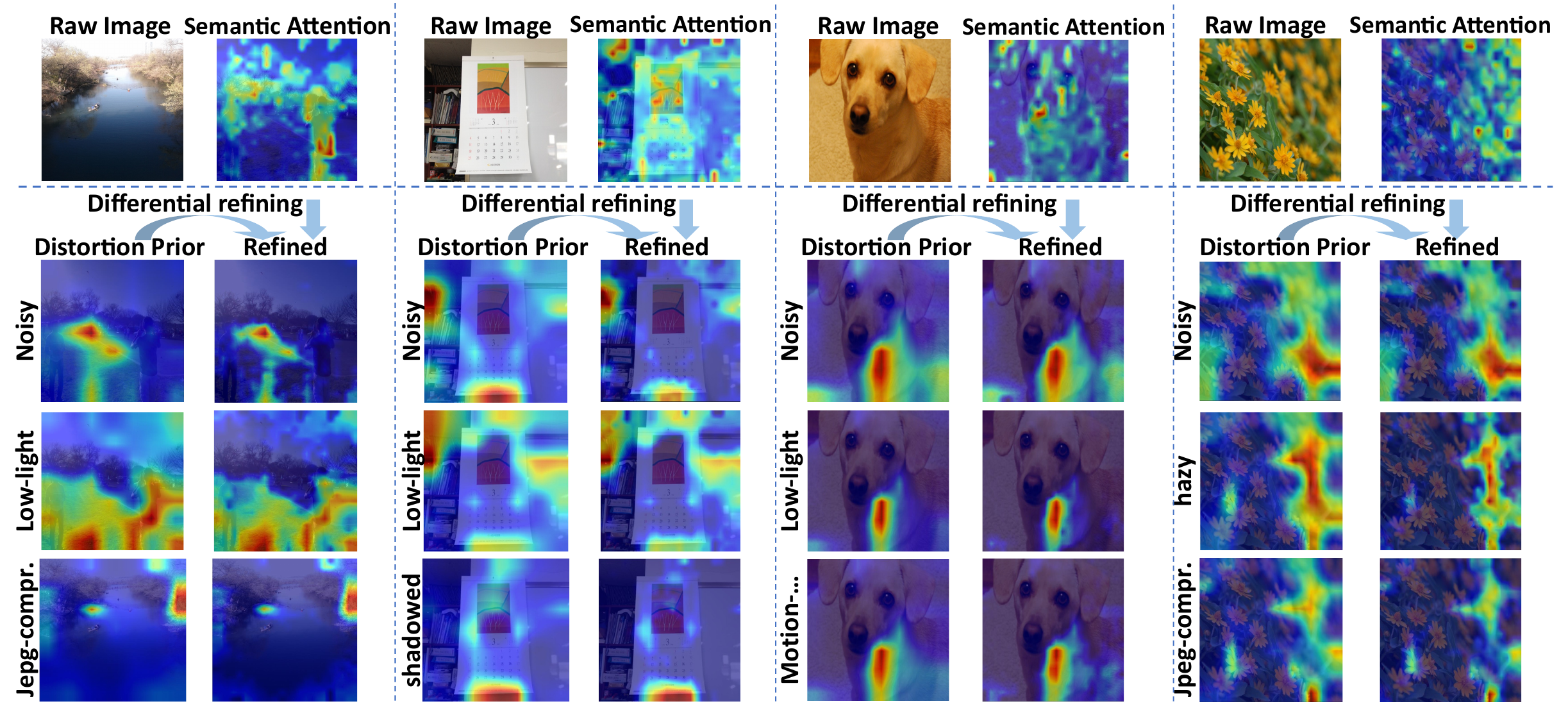}
  \caption{Comparison of attention maps from DA-CLIP, the image encoder, and DSDM outputs.}
  \label{fig:visual}
\end{figure*}

\subsection{Ablation Study}

\paragraph{Module Study}
We conducted ablation experiments to evaluate the effectiveness of the proposed Distortion-Saliency Differential Module and Dynamic Distortion Weighting Module. As shown in Table~\ref{tab:ablation}, the \emph{Image Encoder} represents the performance when using only ViT as the BIQA backbone, with the results indicating that high-level information for quality assessment is insufficient and thus comes with data scarcity, as that the model performance on the smaller dataset LIVEC is less comparable against large dataset. \emph{DA-CLIP} refers to the approach where only the prior module is involved with the vision and language encoder frozen, followed by a dot-product operation identical to that of CLIPIQA to obtain distortion-specific features. We only trained the prediction head with these features to regress scores. The results demonstrate that the features extracted by DA-CLIP are not directly applicable to BIQA, as certain distortion types, \emph{e.g.}, rain-drop and uncompleted, are either irrelevant or detrimental to visual quality. These results are also broadly consistent with the observations in CLIPIQA. Based on these features, \emph{DSDM} is introduced to improve the priors and purify the attentions, thereby mitigating noise interference in BIQA and enhancing the final performance, especially on smaller datasets. Furthermore, \emph{DR.Experts} means the inclusion of the proposed DDWM module, which leverages a comprehensive feature group to differentiate and optimize the contributions of various distortions, \emph{i.e.}, levels and types, to quality scores, further enhancing the metrics on both the KonIQ and LIVEC datasets.

\begin{table}[!ht]
    \centering
    % \resizebox{0.9\columnwidth}{!}{
    \begin{tabular}{lcccc}
        \toprule[1.5pt]
         & \multicolumn{2}{c}{KonIQ} & \multicolumn{2}{c}{LIVEC} \\
        \cmidrule(lr){2-3} \cmidrule(lr){4-5}
         Methods & SRCC & PLCC & SRCC & PLCC \\
        \midrule
        Image Encoder & 0.916 & 0.929 & 0.857 & 0.884 \\
        DA-CLIP & 0.720 & 0.754 & 0.587 & 0.635 \\
        DSDM & 0.930 & 0.941 & 0.885 & 0.904 \\
        % DF + DDWM[quer+uer] & 0.939 & 0.950 & 0.899 & 0.916 \\
        % DF + DDWM[query]  & -- & -- & 0.898 & 0.913 \\
        \midrule[1pt]
        DR.Experts  & \textbf{0.941} & \textbf{0.954} & \textbf{0.914} & \textbf{0.926} \\
        \bottomrule[1.5pt]
    \end{tabular}
    % }
    \caption{Ablation experiments on LIVEC and KonIQ datasets. Bold entries indicate the best performance.}
    \label{tab:ablation}
\end{table}

\paragraph{Feature Group Study}
Table~\ref{tab:Comprehensive Feature Group} compares different compositions of the feature group in DDWM. Specifically, \emph{Dis}, \emph{Sem}, and \emph{Bri} represent the distortion features refined by DSDM, the semantic features extracted by the image encoder, and the supplementary bridging features derived from the difference between the two aforementioned feature groups, respectively. To note, \emph{Only} indicates that only a specific feature group is included, while \emph{w/o} suggests the inclusion of the other two feature groups yet excluding the current one. The results demonstrate that as more feature groups are introduced, the proposed method encompasses more information, resulting in more comprehensive weighing and better reflecting the quality score of the current image.

\begin{table}[ht]
    \centering
    % \resizebox{0.9\columnwidth}{!}{
    \begin{tabular}{lcccc}
        \toprule[1.5pt]
         & \multicolumn{2}{c}{KonIQ} & \multicolumn{2}{c}{LIVEC} \\
        \cmidrule(lr){2-3} \cmidrule(lr){4-5}
        Methods & SRCC & PLCC & SRCC & PLCC \\
        \midrule
        Only Dis   & 0.933 & 0.944 & 0.897 & 0.912 \\
        Only Sem    & 0.936 & 0.944 & 0.899 & 0.913 \\
        Only Bri & 0.936 & 0.947 & 0.897 & 0.914 \\
        \midrule
        w/o Dis  & 0.938 & 0.948 & 0.913 & 0.918 \\
        w/o Sem   & 0.940 & 0.951 & 0.911 & 0.919 \\
        w/o Bri & 0.939 & 0.950  & 0.909 & 0.916 \\
        \midrule[1pt]
        Full(Ours)  & \textbf{0.941} & \textbf{0.954} & \textbf{0.914} & \textbf{0.926} \\
        \bottomrule[1.5pt]
    \end{tabular}
    % }
    \caption{Ablation study on different compositions of feature groups. Bold entries indicate the best performance.}
    \label{tab:Comprehensive Feature Group}
\end{table}

\paragraph{Qualitative Analysis}
Fig.~\ref{fig:visual} illustrates a comparative analysis involving three types of attention maps: attention maps on specific distortion obtained from DA-CLIP, semantic attention maps produced by the image encoder, and refined attention maps on distortions through DSDM. Technically, we visualize only the top three distortions, identified by the DDWM, of the most significance on image quality. From the figure, it can be observed that the refined attention maps effectively suppress irrelevant areas related to semantic information, while easing attention noise introduced by fictitious distortions. These results indicate that the proposed DSDM significantly enhances the ability to capture and refine important distortion features in BIQA tasks.

\section{Conclusion}
We propose DR.Experts, a BIQA framework leveraging distortion priors refined via a distortion-aware vision-language model. Its Dynamic Distortion Weighting Module adaptively weights features using cues like distortion characteristics and semantics to simulate human vision. This design reduces over-reliance on unified features and enhances data efficiency. Extensive experiments confirm DR.Experts’ superior performance and generalization across BIQA benchmarks.

\section{Acknowledgments}
This work was supported in part by the National Science Foundation of China (Grant No. 62301041), and in part by Beijing Institute of Technology Research Fund Program for Young Scholars.

\bibliography{aaai2026}

\end{document}